%% file: mlinearDP.tex
\documentclass[10pt, conference, compsocconf, onecolumn]{IEEEtran}

\usepackage{textgreek}
\usepackage{amsfonts,amsmath}
\usepackage{amsthm}
\usepackage{algpseudocode}
\usepackage{graphicx}
\usepackage{balance}
\usepackage{algorithm}
\usepackage{tabularx,ragged2e,booktabs,caption}
\usepackage[mathscr]{euscript}
\usepackage{amsbsy}
\usepackage{mathtools}
\usepackage{notes}
\usepackage[utf8]{inputenc} 
\usepackage[T1]{fontenc}    
\usepackage{hyperref}       
\usepackage{url}            
\usepackage{nicefrac}       
\usepackage{microtype}      

\newcommand{\upperRomannumeral}[1]{\uppercase\expandafter{\romannumeral#1}}

\theoremstyle{definition}
\newtheorem{definition}{Definition}

\begin{document}

\title{Multilinear Dirichlet Processes}
\author{Xiaoli Li\\
xiaoli.li@ainstein.ai \\
Ainstein AI
}
\date{}

\maketitle
\input{abstract}
%

\input{introduction}
\input{related-work}
\input{preliminary}

\input{method}
\input{experiment}
\input{conclusion}
\bibliographystyle{abbrv}
\small{
\bibliography{mlinearDP}
}
\end{document}

%% file: abstract.tex
\begin{abstract}
Dependent Dirichlet processes (DDP) have been widely applied to model data from distributions over collections of measures which are correlated in some way. On the other hand, in recent years, increasing research efforts in machine learning and data mining have been dedicated to dealing with data involving interactions from two or more factors. However, few researchers have addressed the heterogeneous relationship in data brought by modulation of multiple factors using techniques of DDP. In this paper, we propose a novel technique, MultiLinear Dirichlet Processes (MLDP), to constructing DDPs by combining DP with a state-of-the-art factor analysis technique, multilinear factor analyzers (MLFA). We have evaluated MLDP on real-word data sets for different applications and have achieved state-of-the-art performance.
\end{abstract} 

%% file: introduction.tex
\section{Introduction}
Dependent Dirichlet processes (DDP) have been widely applied to model data from distributions over collections of measures which are correlated in some way. To introduce dependency into DDP, various techniques have been developed via correlating through components of atomic measures, such as atom sizes \cite{dunson08a, griffin06, rodriguez11} and atom locations \cite{de04, gelfand05, teh04}, sampling from a DP with random distributions as atoms \cite{rodriguez08}, operating on underlying compound Poisson processes \cite{lin10}, regulating by L\'evy Copulas \cite{leisen13}, or constructing those measures through a mixture of several independent measures drawn from DPs \cite{hatjispyros16,kolossiatis13, ma15, muller04}.

On the other hand, in recent years, increasing research efforts in machine learning and data mining have been dedicated to dealing with heterogeneously related data involving interactions from two or more factors. For example, in multilinear multi-task learning \cite{romera13}, predictions of a student's achievement may be affected by both her school environment and time. In context aware recommender systems, different conceptual factors, such as time and companions, play major roles on a user's preferences for restaurants.

However, few researchers have addressed the heterogeneous relationship in data brought by modulation of multiple factors using techniques of DDP. To the best of our knowledge, the only work that considered multiple groups of factors was proposed by De Iorio et al. \cite{de09, de04}. In their work, the dependence into collections of related data was introduced by building an ANOVA structure across atom locations of random measures. The main weakness of ANOVA based DDP is that the model becomes cumbersome when the number of factor groups increases and the inference may be computationally daunting, especially when the multiplicative interactions of factors are also included in the ANOVA effects. In addition, the method assumes that the effects of a factor are the same for those samples which are affected by that factor. However, this assumption may be invalid in some situations. For example, the school environment may have varying degrees of impact in the academic performance of each student. 

In this paper, we propose a novel technique of constructing DDP based on DP and multilinear factor analyzers (MLFA) \cite{tang13} to overcome the limitations in aforementioned works. We refer to this method as Multilinear Dirichlet Processes (MLDP). Specifically, we are trying to model $S$ sets of samples that are correlated through $N$ groups of factors by constructing $S$ dependent random measures, with one random measure used to model the distribution of one set of samples. To capture the correlations among different sets of samples, we hypothesize that those $S$ dependent random measures are the results of multiplicative interactions of $N$ groups of factors. Specifically, we represent each random measure as a linear combination of $I$ different latent basis measures. We may consider each basis measure as a 1-mode fiber of a shared latent factor tensor in MLFA. Then we determine the weights of those linear combinations using multiplicative interactions of latent parameter vectors that correspond to different factor groups by borrowing the ideas from MLFA.

To evaluate the performance of MLDP, we have compared it with DP-based methods using $4$ real world data sets. In addition, we have applied MLDP to various machine learning problems, multilinear multi-task learning, and context-aware recommendation, which have received much attention from researchers recently. The comprehensive experiments demonstrate the effectiveness of MLDP on different applications.

The contribution of this work is two-fold:
\begin{itemize}
\item We have developed a novel technique MLDP to construct DDPs by combine DP with multiliear Factory Analyzers to model the distributions of data modulated by different factors.
\item We have demonstrated the effectiveness of MLDP by applying it to two different applications, multilinear multi-task learning and context-aware recommender systems and evaluating MLDP on $4$ real-word data sets. The state-of-the-art performance achieved by MLDP has validated its applicability to those applications.
\end{itemize}

The remainder of the paper is presented in the following way. We first describe related studies in section \ref{sec:relatedwork}. Then we describe some basic definitions about tensors and give a brief description of multilinear factor analyzers in section \ref{sec:preliminary}. In section \ref{sec:algorithm} we present the formulation of MLDP. We describe experimental setup and results in section \ref{sec:experiments}. A detailed discussion is also given in this section. We conclude the whole work in section \ref{sec:conclusion}.

%% file: related-work.tex
\section{Related Work}\label{sec:relatedwork}
Dependent Dirichlet Processes has long standing in the literature of Bayesian nonparametric methods. A wide variety of techniques have been developed to address various correlations of sets of samples. The interested reader is referred to a comprehensive survey conducted by Foti and Williamson \cite{foti15}. Those techniques can be categorized into two groups according to the heterogeneity of underlying factors which a method aims to capture for untangling the correlations. Most of existing techniques of DDP, which belong to the first group, only consider one underlying factor which leads to the correlations among samples, such as space \cite{gelfand05}, time \cite{caron07}, study \cite{muller04}, or pairwise distance \cite{blei11}.

For the methods in the second group, multiple groups of factors were taken into account when modeling correlations. Compared with the considerably large number of previous works in the first group, few studies have been dedicated to capturing multiple factors.  De Iorio et al. \cite{de04} proposed a method to model dependence across related random measures by building an ANOVA dependence structure among atom locations of random measures. In a later work, De Iorio et al. \cite{de09} further proposed a linear DDP model by extending the ANOVA DDP model to include continuous covariates through a DP mixture of linear models.




%% file: preliminary.tex
\section{Preliminary}\label{sec:preliminary}
In this section, we introduce necessary background knowledge about tensors and Multilinear Factor Analyzers on which our method is based.
\subsection{Notations}
For clarity, we introduce the notations that will be used throughout the paper.  We use lowercase letters to represent scalar values, lowercase letters with bold font to represent vectors (e.g. $\vect{u}$), uppercase bold letters to represent matrices (e.g. $\mat{A}$), and Euler script letters to represent Tensors (e.g. $\mathscr{T}$). Unless stated otherwise, all vectors in this paper are column vectors. We use $[1:N]$ to denote the set $\{1,2,\dots,N\}$.
\subsection{Basic definitions for Tensors}
Following \cite{lathauwer00, kolda09}, we introduce some basic definitions for tensors.

\begin{definition}[Order]
 The order of a tensor is defined as the number of dimensions of a tensor. A Nth-order tensor, denoted as $\mathscr{A} \in \mathbb{R}^{I_1\times I_2\times\cdots\times I_n}$, has $N$ indices with each index addressing a mode of $\mathscr{A}$.
\end{definition}

\begin{definition}[n-mode Fiber]
Fibers are defined as vectors that constitute a tensor. A n-mode tensor fiber is obtained by varying the $nth$ index of the tensor while fixing all other indices. Columns and rows are 1-mode and 2-mode fibers for a matrix, i.e. a 2nd-order tensor, respectively.
\end{definition}

\subsection{Multilinear Factor Analyzers}
As an extension of bilinear models  \cite{tenenbaum00}, which was originally developed to untangle ``content" and ``style" factors in images,  Multilinear factor analyzers (MLFA) \cite{tang13} were developed to model data that is the result of multiplicative interactions of $N$ groups of factors, with $J_n$ factors in each group. Let denote the $j_n$th factor in factor group $n$ using a latent parameter vector $\vect{z}^{n,j_n}$, where $j \in [1:J_n]$,  $\vect{z}^{n,j_n} \in \reals^{I_n}$, and $n \in [1:N]$,  then MLFA formulate an observed vector $\vect{x}^{j_1,\dots,j_N} \in \mathbb{R}^P$ modulated by factors $j_1,\dots,j_N$ using a shared latent factor tensor $\mathscr{D} \in \mathbb{R}^{P\times I_1\times I_2\times\dots\times I_N}$ as follows \cite{li16}:
\begin{align} \label{eq:tensorVectorEM}
\vect{x}^{j_1,\dots,j_N} = \sum\limits_{i_1,\dots, i_N} ( \prod\limits_{n=1}^N z^{n,j_n}_{i_n} )  \vect{d}_{:,i_1,\dots, i_N} + \epsilon
\end{align}
where $\vect{d}_{:,i_1,\dots, i_N}$ is a 1-mode fiber of $\mathscr{D}$. $z^{n,j_n}_{i_n}$ is the $i_n$th element of the vector $\vect{z}^{n,j_n}$. $\epsilon$ is an i.i.d error term following a multivariate Normal distribution.


%% file: method.tex
\section{Algorithm}\label{sec:algorithm}
In this section ,we first formulate the problem we aim to solve. Then we describe the details of the proposed method multilienar Dirichlet Processes (MLDP). Lastly, we present the inference algorithm we have developed for MLDP.

\subsection{Problem Formulation}
Given $N$ groups of factors, and $J_n$ observed factors in the $n$th group , we collect a set of samples, $\mat{X}^{j_1,\dots,j_N} \in \reals^{M^{j_1,\dots,j_N} \times P}$, for each combination of factors from $N$ groups, where one factor is used for each group, to get $S = J_1\times J_2 \dots J_N$ sets of samples in total. Here $j_n$ is used to specify that $j_n$th factor in group $n$ is used in the combination. For example, suppose there are $2$ groups of factors with $2$ factors in each group, i.e. $J_1=J_2=2$, then we have $S = J_1 \times J_2 =4$ sets of samples $\mat{X}^{1,1}$, $\mat{X}^{2,1}$, $\mat{X}^{1,2}$, $\mat{X}^{2,2}$, where $\mat{X}^{1,2}$ is affected by two factors, 1st factor in factor group $1$ and 2nd factor in factor group $2$.

Our goal is to fit $S$ sets of samples using generative models based on DDP techniques. Note that those $S$ sets of samples are modulated by the interactions of $N$ groups of factors. Therefore, to better fit the data, a model that can capture the interactions among different factors is needed. To this end, we propose Multilinear Dirichlet Processes by borrowing ideas of modeling interactions among factors from Mulitlinear factor analyzers (MLFA).  Note that MLFA were designed for factor analysis of observed samples represented in vectors. It cannot be directly applied to modeling distributions. To leverage the advantage of MLFA for tackling multiplicative interactions of factors, we equate a group of basis random measures with the shared latent factor tensor $\mathscr{D}$ in MLFA, treating each basis measure as a 1-mode fiber of $\mathscr{D}$. Then a random measure can be constructed using a linear combination of those basis random measures by determining the weights of linear combinations using the same technique in MLFA. However, there is another challenge posed by employing MLFA for model random distribution.  In MLFA, the weights of linear combinations can be any real numbers. In order to construct a valid random measure, the weights of a linear combination must be positive and sum to one. To this end, we utilize a softmax function to normalize the weights.

Before proceeding to the formulation of MLDP, we summarize important notations for MLDP in Table \ref{tb:notations}.

\begin{table}
\centering
\caption{Notations for MLDP}\label{tb:notations}
\begin{tabular}{|r|l|}
  \hline
  $N$       &  Total number of factor groups\\
  $J_n$         & Total number of factors in the $n$th factor group\\
  $S$       & Total number of sets of samples \\
  $\mat{X}^{j_1,\dots,j_N}$ & The set of samples modulated by factors $j_1,\dots,j_N$ \\
  $I_n$ & Total number of basis measures for the $n$th factor group\\
  $G^{*}_{i_1,\dots,i_N}$ &  The basis measure indexed by $(i_1,\dots,i_N)$ \\			
  $G^{j_1,\dots,j_N}$ & The random measure used to model $\mat{X}^{j_1,\dots,j_N}$\\
  $\vect{u}^{n,j_n}$ &  The latent parameter vector for $j_n$th factor in factor group $n$\\
  $w_{i_1,\dots,i_N}^{j_1,\dots,j_N}$ & The linear combination weight of $G^{*}_{i_1,\dots,i_N}$ for $G^{j_1,\dots,j_N}$\\
  \hline
\end{tabular}
\end{table}
\subsection{Multilinear Dirichlet Processes}

Given $N$ groups of factors, we assume that each factor in a factor group corresponds to a latent parameter vector $\vect{u} \in \reals^{I_n}$. Then we use linear combinations of $I = I_1\times I_2\dots I_N$ basis measures $G^*$ to define Multilinear Dirichlet Processes (MLDP) as follows:
\begin{align}\label{eq:mldp}
G^{j_1,\dots,j_N} = \sum\limits_{i_1,\dots,i_N}w_{i_1,\dots,i_N}^{j_1,\dots,j_N}G^{*}_{i_1,\dots,i_N}  \nonumber\\
G^*_{i_1,\dots,i_N} \sim DP(\alpha, H) \nonumber \\
for \; j_n \in [1:J_n], i_n \in [1:I_n], n\in [1:N]
\end{align}
$G^*_{i_1,\dots,i_N}$ is represented using the form:
\begin{align}
G^*_{i_1,\dots,i_N} = \sum\limits_{k=1}^{\infty}\pi^{k}_{i_1,\dots,i_N}\delta_{\phi^k_{i_1,\dots,i_N}} \nonumber
\end{align}
Where the weights $\pi^{k}_{i_1,\dots,i_N}$ can be iteratively constructed using a stick-breaking process with parameter $\alpha$  \cite{sethuraman94}. And each atom $\phi^k_{i_1,\dots,i_N}$ is an i.i.d draw from base distribution $H$.

The weights for the linear combinations of basis measures $G^*$'s are determined by latent parameter vectors $\vect{u}$'s through softmax functions:
\begin{align}
 w_{i_1,\dots,i_N}^{j_1,\dots,j_N} = \frac{e^{u_{i_1}^{1,j_1}u_{i_2}^{2,j_2}\dots u_{i_N}^{N,j_N}}}
{\sum\limits_{k_1,k_2,\dots,k_N}e^{u_{k_1}^{1,j_1}u_{k_2}^{1,j_2}\dots u_{k_N}^{N,j_N}}} \quad \nonumber \\
u^{n,j_n}_{i_n} \sim \textup{N}(0,(\sigma_u^n)^2)  \quad  \log((\sigma_u^n)^2) \sim \textup{N}(0, \sigma_0^2)  \nonumber \\
for \; i_n \in [1:I_n], j_n \in [1:J_n], n \in [1:N] \nonumber
\end{align}
Where $\vect{u}^{n,j_n}=[u^{n,j_n}_{1},\dots, u^{n,j_n}_{I_n}]^T$ is a latent parameter vector for $j_n$th factor in factor group $n$. $u_{i_n}^{n,j_n}$ is the $i_n$th element of $\vect{u}^{n,j_n}$. Note that $w_{i_1,\dots,i_N}^{j_1,\dots,j_N}$ has the property that $\sum\limits_{i_1,\dots,i_N}w_{i_1,\dots,i_N}^{j_1,\dots,j_N} = 1$

\textbf{Properties of MLDP.} Let denote all the latent parameter vectors $\vect{u}$'s as $\mat{U}$, then it is apparent that for any Borel set $B$, we have:
\begin{align}
E\{G^{j_1,\dots,j_N}(B)|\mat{U}\} = \sum\limits_{i_1,\dots,i_N}w_{i_1,\dots,i_N}^{j_1,\dots,j_N} H(B) = H(B) \nonumber \\
V\{G^{j_1,\dots,j_N}(B)|\mat{U}\} = \sum\limits_{i_1,\dots,i_N}\frac{(w_{i_1,\dots,i_N}^{j_1,\dots,j_N})^2}{1+\alpha} H(B)(1-H(B)) \nonumber
\end{align}
From the above properties of MLDP, we can see that the expectation of each random distribution, $G^{j_1,\dots,j_N}(B)$, which corresponds to a combination of factors from $N$ groups, is the same given $\mat{U}$. And the difference in variance of $G^{j_1,\dots,j_N}(B)$ is determined by $\sum\limits_{i_1,\dots,i_N}(w_{i_1,\dots,i_N}^{j_1,\dots,j_N})^2$.

It is worth noting that DP is a special case of MLDP when the dimensions of $\vect{u}$'s are $1$.
\begin{align}
 w_{1,1,\dots,1}^{j_1,\dots,j_N} = \frac{e^{u_{1}^{1,j_1}u_{1}^{2,j_2}\dots u_{1}^{N,j_N}}}
{e^{u_{1}^{1,j_1}u_{1}^{1,j_2}\dots u_{1}^{N,j_N}}} =1 \nonumber \\
G^{j_1,\dots,j_N} = w_{1,1,\dots,1}^{j_1,\dots,j_N}G^{*}_{1,1,\dots,1} = G^{*} \nonumber \\
for \; j_n \in [1:J_n], n\in [1:N] \nonumber
\end{align}
From the above derivation, we can see that here is only one basis measure when the dimensions of $\vect{u}$'s are $1$ and this basis measure is a draw from a DP. That is, $G^{j_1,\dots,j_N}$ is a draw from a classical DP and a MLDP degenerates to a DP. 

Note that in the above definition of MLDP \ref{eq:mldp}, we assume that basis measures are drawn from the same $DP(\alpha, H)$ to allow rather limited heterogeneity in data. On the other end of the spectrum of the heterogeneity,  we may use DP's with different parameters for $G^*$s to have:
\begin{align} \label{eq:mmldp}
G^*_{i_1,\dots,i_N} \sim DP(\alpha_{i_1,\dots,i_N}, H_{i_1,\dots,i_N}) \quad for \; i_n \in [1:I_n], n \in [1:N]
\end{align}
$G^*_{i_1,\dots,i_N}$ can be represented using the form:
\begin{align}
G^*_{i_1,\dots,i_N} = \sum\limits_{k=1}^{\infty}\pi^{k}_{i_1,\dots,i_N}\delta_{\phi^k_{i_1,\dots,i_N}} \nonumber
\end{align}
Similarly, the weights $\pi^{k}_{i_1,\dots,i_N}$ can be iteratively constructed using a stick-breaking process with parameter $\alpha_{i_1,\dots,i_N}$. And each atom $\phi^k_{i_1,\dots,i_N}$ is an i.i.d draw from base distribution $H_{i_1,\dots,i_N}$.

Although the high flexibility of MLDP allows it to model extremely heterogeneous data, this may entail the issue of unidentifiability. To address this issue, we may add constraints to those factor loadings $\vect{u}$'s or induce sparsity into MLDP. Specially, we may consider using sparsity-promotion priors, such as a hierarchical Student-t prior \cite{tipping01} or a spike-and-slab prior \cite{ishwaran05}, to allow a small set of basis measures are used and improve identifiability. We defer this topic to future work.



\subsection{MLDP Mixture of Models}
Having defined MLDP, the mixture of models using MLDP is straightforward. Given a set of samples, $\mat{X}^{j_1,\dots,j_N} \in \reals^{M^{j_1,\dots,j_N} \times P}$, for each combination of factors from $N$ groups, where there are $J_n$ factors in the $nth$ factor group. We use the following generative model for $\mat{X}$'s
\begin{align}
\vect{x}_m^{j_1,\dots,j_N} \sim f(\cdot|\theta_m^{j_1,\dots,j_N})   \nonumber \\
\theta_m^{j_1,\dots,j_N} \sim G^{j_1,\dots,j_N}  \nonumber  \\
G^{j_1,\dots,j_N} \sim MLDP  \nonumber \\
for \; m \in [1:M^{j_1,\dots,j_N}], j_n \in [1:J_n], n\in [1:N] \nonumber
\end{align}
where $\vect{x}_m^{j_1,\dots,j_N} \in \reals^{1 \times P}$ is the $m$th row of $\mat{X}^{j_1,\dots,j_N}$.

\subsection{Computation}
We use approximate inference based on Markov Chain Monte Carlo (MCMC) methods for MLDP since the inference cannot be obtained analytically. Specifically, we use a Gibbs sampler to approximate the posterior distribution of model parameters $(\phi^k_{i_1,\dots,i_N}, \vect{u}^{n,j_n}, \sigma_u^n)$ by extending Algorithm 8 proposed in \cite{neal00} since it can handle non-conjugate base measures by using auxiliary clusters.

The major difference between the inference of MLDP and that of classical DP is that we also need to determine which basis measure, i.e. $G^{*}$, is used for a specific sample in addition to cluster assignment decisions. Thus we describe here only steps about how to assign a sample to a basis measure and a cluster of that basis measure due to space limit.  The updating of $\vect{u}^{n,j_n}$'s, $\sigma_u^n$'s, and $\phi^k_{i_1,\dots,i_N}$'s is straightforward after we have decided cluster and basis measure assignments.

We first introduce additional indicator variables $b_m^{j_1,\dots,j_N}$ to specify which basis measure is used for a sample. To be specific, we have $b_m^{j_1,\dots,j_N} = (i_1,\dots,i_N)$ if and only if the corresponding basis measure for $\vect{x}_m^{j_1,\dots,j_N}$ is $G^{*}_{i_1,\dots,i_N}$.
In addition, similar to the inference in classical DP, we also introduce a latent indicator variable $c_m^{j_1,\dots,j_N}$, which specify the cluster a sample $\vect{x}_m^{j_1,\dots,j_N}$ belongs to, to facilitate the inference. We use the following procedure in each iteration to update $b_m^{j_1,\dots,j_N}$, $c_m^{j_1,\dots,j_N}$, for $ m \in [1:M^{j_1,\dots,j_N}]$, $j_n \in [1:J_n]$, $n\in [1:N]$.

First let define:
\begin{align}
\rho_{i_1,\dots,i_N}^k = \begin{cases}
r\frac{w_{i_1,\dots,i_N}^{j_1,\dots,j_N}l_{i_1,\dots,i_N}^{-m,k}}{L_{i_1,\dots,i_N}^{-m}+\alpha}f(\vect{x}_m^{j_1,\dots,j_N}|\phi^k_{i_1,\dots,i_N}) \quad for \; k=1,\dots,K_{i_1,\dots,i_N}^{-m}  \\
 r\frac{w_{i_1,\dots,i_N}^{j_1,\dots,j_N}\alpha/s}{L_{i_1,\dots,i_N}^{-m}+\alpha}f(\vect{x}_m^{j_1,\dots,j_N}|\phi^k_{i_1,\dots,i_N}) \quad for \; k=K_{i_1,\dots,i_N}^{-m}+1,\dots,K_{i_1,\dots,i_N}^{-m}+s \nonumber \\
 \end{cases}
\end{align}
here $r$ is the appropriate normalizing constant; $s$ is the number of auxiliary clusters; $K_{i_1,\dots,i_N}^{-m}$ is the number of active clusters in basis measure $G^*_{i_1,\dots,i_N}$; $L_{i_1,\dots,i_N}^{-m}$ is the total number of samples assigned to the basis measure $G^{*}_{i_1,\dots,i_N}$; and $l_{i_1,\dots,i_N}^{-m,k}$ is the number of samples that are allocated to cluster $k$. Note that we use the superscript $-m$ to denote that the sample $\vect{x}_m^{j_1,\dots,j_N}$ is excluded. $\phi^k_{i_1,\dots,i_N}$'s are drawn from the base distribution of $G^{*}_{i_1,\dots,i_N}$ when $K_{i_1,\dots,i_N}^{-m}+1\leq k \leq K_{i_1,\dots,i_N}^{-m}+s$.

Then we generate a draw $(b_m^{j_1,\dots,j_N}$, $c_m^{j_1,\dots,j_N})$ based on the following probability:
\begin{align}
p(b_m^{j_1,\dots,j_N}=i_1,\dots,i_N, c_m^{j_1,\dots,j_N}=k | \vect{x}_m^{j_1,\dots,j_N},\mat{U}, B^{-m}, C^{-m}, \Phi) = \rho_{i_1,\dots,i_N}^k
\end{align}
where we use $\mat{U}$ to denote all the latent vectors $\vect{u}^{n,j_n}$'s. $B^{-m}$ and $C^{-m}$ are used to denote sets of indicator variables $b_m^{j_1,\dots,j_N}$'s and $c_m^{j_1,\dots,j_N}$'s respectively without considering $\vect{x}_m^{j_1,\dots,j_N}$. And $\Phi$ is the set of $\phi^k_{i_1,\dots,i_N}$'s.


%% file: experiment.tex
\section{Experiments}\label{sec:experiments}
In this section, we evaluate the performance of MLDP by applying it to two different applications: multilinear multi-task learning (MLMTL), and context-aware recommendation system (CARS) using $4$ real-world data sets. For each data set, we randomly selected $50\%$ of samples as training data set and used all the rest as test data set. 
We repeated this process $10$ times for each data set and reported the averaged performance on the test data set.

\subsection{Multilinear Multi-task Learning (MLMTL)}
In multilinear multi-task learning (MLMTL), each task is associated with two or multiple modes. For example, in predicting ratings given by a specific consumer to different aspects of a restaurant, such as food quality or service quality, a MLMTL algorithm formulates the problem by considering each combination of one consumer and one aspect (two modes) as a task using a 2-dimensional indexing.

To handle MTMTL, we use MLDP Mixture of Regression Models (MLDP-MRM) by treating each mode as a factor group. Suppose there is a set of tasks associated with $N$ modes with $J_n$ aspects in the $nth$ mode. For a task indexed by $(j_1,\dots,j_N)$, we obtain a set of samples $(\mat{X}^{j_1,\dots,j_N}, \vect{y}^{j_1,\dots,j_N})$, where $\mat{X}^{j_1,\dots,j_N} \in \reals^{M^{j_1,\dots,j_N} \times P}$ and $\vect{y}^{j_1,\dots,j_N} \in \reals^{M^{j_1,\dots,j_N}}$. The following MLDP-MRM model is used in our experiments:
\begin{align}
\vect{x}_m^{j_1,\dots,j_N} \sim \textup{N}(\vect{\mu}_{x,m}^{j_1,\dots,j_N}, \mat{\Sigma}_{x,m}^{j_1,\dots,j_N})  \nonumber \\
y_m^{j_1,\dots,j_N} \sim \textup{N}(\vect{x}_m^{j_1,\dots,j_N}\vect{\beta}_m^{j_1,\dots,j_N}, \sigma_{y,m}^{j_1,\dots,j_N})  \nonumber \\
(\vect{\mu}_m^{j_1,\dots,j_N}, \mat{\Sigma}_{x,m}^{j_1,\dots,j_N}, \vect{\beta}_m^{j_1,\dots,j_N}, \sigma_{y,m}^{j_1,\dots,j_N}) \sim G^{j_1,\dots,j_N} \nonumber \\
G^{j_1,\dots,j_N} \sim MLDP  \nonumber  \\
for \; m \in [1:M^{j_1,\dots,j_N}], j_n \in [1:J_n], n\in [1:N] \nonumber
\end{align}
For computationally convenient, we use the following priors for basis distribution $H$
\begin{align}
H(\vect{\mu}_x, \mat{\Sigma}_{x}, \vect{\beta}, \sigma_y)=\textup{NIW}(\vect{\mu}_x, \mat{\Sigma}_{x}; \vect{\mu}^0_x, \lambda_0, \mat{\Psi}^0_x, \nu_0) \textup{N}(\vect{\beta};\vect{0}, \sigma_y^2\mat{V})\textup{IG}(\sigma_y^2; a_y, b_y)
\end{align}

We compared MLDP-MRM with $3$ other Mixture of Regression Models, DP-MRM, MXDP-MRM and ANOVADP-MRM, using different mixing measure, DP, MXDP \cite{muller04}, and ANOVADP \cite{de04} respectively. In addition, we also used two state-of-the-art MLMTL algorithms MLMTL-C, which is based on convex tensor trace norm regularization(\cite{romera13}), and TPG \cite{yu16}, which is based on prototypical method of projected gradient descent, for comparison.  
$2$ real-world data sets, restaurant data set and school data set, were utilized for the experiments, which we describe in the following.

\textbf{Restaurant \& Consumer Data Set}. This data set contains $1161$ ratings, including food rating, service rating, and overall rating, from $131$ consumers for $130$ restaurants \cite{vargas11}. The task is to predict a consumer's rating for a restaurant given the attributes of the consumer and the restaurant. 
We converted categorical attributes using binary coding to obtain $71$ features for each sample. Then we applied PCA to the training data set to keep the first $25$ components and then performed the same transformation on the test data set using the learned loadings. There are $2$ groups of factors, corresponding to consumers and different aspects of the ratings. For the number of factors, we have $J_1 = 131$ and $J_2 = 3$.

\textbf{School Data Set}. The school data set consists of examination records from 140 secondary schools in years 1985,1986 and 1987. The attributes of the data include the year of examination, 4 school-specific attributes, and 3 student-specific attributes, where categorical attributes are expressed as binary features \cite{argyriou08}. The number of features used in the experiments were $19$ after applying PCA. We organized the data according to $2$ groups of factors, corresponding to schools and years of examination. We excluded those schools which did not contain records from all $3$ years to obtain $64$ schools. Thus we have $J_1=64$ and $J_2=3$. The prediction goal is to estimate a student's examination score.

Root mean squared error (RMSE) was employed to evaluate the results. The performance of different algorithms is showed in the first $2$ sub-figures of Fig. \ref{fig:realPerf}. On the restaurant \& consumer data set, we observe that DDP-based methods, i.e. MXDP-MRM, ANOVADP-MRM, and MLDP-MRM, outperforms DP-MRM with a large margin. Compared with MXDP-MRM and ANOVADP-MRM, MLMTL-C, which is specially designed for MLMTL, has a clear advantage. However, it is worth noting that our proposed method has achieved better performance than both MLMTL and TPG. The results on school data set, showed in the 2nd sub-figure of Fig. \ref{fig:realPerf}, present the similar trend except that ANOVADP-MRM performed worse than DP-MRM. This demonstrated the applicability of MLDP to multilinear multi-task learning problems.

\subsection{Context-aware Recommendation}
In context-aware recommender systems, conceptual variables are also considered in making recommendations in addition to the attributes of users and items. In this experiment, we evaluate the performance of rating predictions based on MLDP models. To apply MLDP to context-aware recommendation, we map each conceptual variable to a factor group and treat each context condition as a factor.

Similar to the experiment for MLMTL, we use MLDP-MRM for prediction tasks and compare it with DP-MRM, MXDP-MRM, and ANOVADP-MRM. Furthermore, we compare MLDP-MRM with a context-aware recommender system, CSLIM \cite{zheng14}, to investigate whether MLDP-MRM is competitive with the current state-of-the-art technique. Two real-world data sets, Japan Restaurant Data set and Frappe Data set were utilized in the experiment. We describe them in the following.

\textbf{Japan Restaurant Data Set}. It consists of $800$ ratings from $8$ users for $69$ restaurants in Japan \cite{oku06}. There are $31$ features in the data set. 
The prediction task for this data set is to estimate a user's rating for a restaurant given the restaurant attributes and context conditions. For MLDP, we use $2$ event parameters, holiday and relation, as $2$ factor groups with $J_1=6$ and $J_2=6$. For CSLIM, we use all the features as context parameters.

\textbf{Frappe Data Set}. This data set consists of usage history logs of $4082$ context-aware mobile apps from $957$ users \cite{frappe15}. There are $96203$ entries in total. We randomly selected $2000$ entries for our experiment. 
For MLDP, we use $2$ features, daytime and isweekend, as $2$ factor groups, with $J_1=2$ and $J_2=7$, to organize the data. For CSLIM, we use all the features as context parameter. The prediction goal is to estimate the number of times an app is used by a user.

\begin{figure*}
  \centering
    \includegraphics*[scale=0.45]{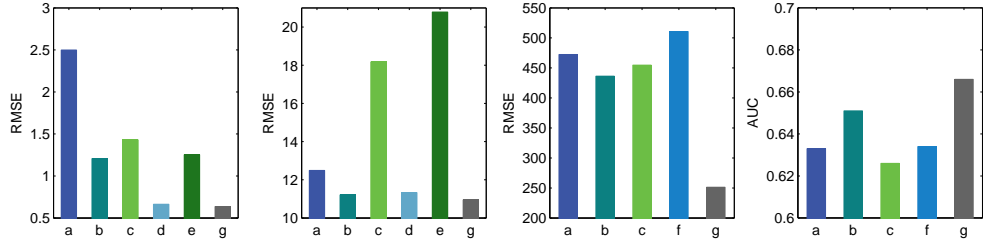}
  \caption{Form Left to Right: 1. MLMTL on Restaurant \& Consumer Data set; 2. MLMTL on School Data set; 3. CARS on Frappe Data Set; 4. CARS on Japan Restaurant Data Set.  Algorithms: a. DP-MRM; b: MXDP-MRM; c: ANOVADP-MRM; d: MLMTL-C; e:TPG; f: CSLIM; g:MLDP-MRM }
  \label{fig:realPerf}
\end{figure*}

We show the results of comparison using the last $2$ sub-figures of Fig. \ref{fig:realPerf}. Similar to MLMTL, RMSE was used for performance evaluation on Frappe data set. For this data set, we observed large variance. The app usage count varies from $1$ to about $20,00$. For the results, we notice that DP or DDP based methods outperforms the state-of-the-art context aware recommender method, CSLIM. Among DP or DDP based methods, our proposed methods MLDP-MRM is significantly better than other methods. It is worth pointing out that we conducted two studies use two different definitions of MLDP, (\ref{eq:mldp}) and (\ref{eq:mmldp}), due to the large variance in the data. We found that MLDP-MRM achieved better performance when using (\ref{eq:mmldp}) (In Fig. \ref{fig:realPerf}, we only show the results of using (\ref{eq:mmldp})). This provides further evidence that MLDP has advantage in handling heterogeneous data.  For Japan restaurant data set, we used AUC since the labels are binary. On this data set, CSLIM performed slightly better than DP-MRM and ANOVADP-MRM while the performance is worse than MXDP-MRM and MLDP-MRM. For MXDP-MRM and MLDP-MRM, the latter performed consistently better.

%% file: conclusion.tex
\section{Conclusion}\label{sec:conclusion}
In this paper, we have devised a novel DDP technique, MLDP, for tackling heterogeneous data modulated by multiple groups of factors, which has been largely ignored in the field of DDP-based methods. To demonstrate the effectiveness of our proposed method, we have applied MLDP to different applications,  multilinear multi-task learning, and context aware recommendations using $4$ real-world data sets. Compared with other state-of-the-art methods, MLDP has achieved better or competitive performance. This confirms the usefulness of MLDP as a way to handling data affected by multiple factors.